\documentclass[10pt]{article}
\usepackage[legalpaper, margin=1in]{geometry}

\usepackage{graphicx}

\usepackage{tikz}
\usepackage{comment}
\usepackage{amsmath,amssymb} 
\usepackage{wrapfig}
\usepackage{hyperref}
\usepackage{amsthm}
\usepackage{color}
\usepackage[algo2e,linesnumbered,ruled]{algorithm2e}
\SetKwComment{Comment}{// }{}
\usepackage{xcolor} 
\usepackage{graphicx}
\usepackage{mathtools}
\usepackage{bm,bbding}
\usepackage{xspace}
\usepackage{subcaption}

\def\Bernoulli{\mathrm{Bernoulli}}
\def\CC{\mathcal{C}}
\def\D{\mathcal{D}}
\def\E{\mathbb{E}}
\def\F{\mathcal{F}}

\def\R{\mathbb{R}}

\def\argmax{\mathop{\rm \arg\,max}\limits}

\newcommand{\RNum}[1]{\uppercase\expandafter{\romannumeral #1\relax}}

\newcommand{\be}{\begin{equationarray}}
\newcommand{\laura}[1]{\bgroup\color{blue}#1\egroup}

\usepackage[pdf]{graphviz}

\usepackage{tkz-graph,tkz-berge}
\usetikzlibrary{arrows,shapes}
\usepackage{mathtools} 
\usepackage{booktabs} 

\usepackage[accsupp]{axessibility}  

\usepackage{booktabs}
\usepackage{enumitem}
\usepackage{subcaption}
\usepackage{bbm}
\usepackage{multirow}

\newtheorem{definition}{Definition}
\newtheorem{theorem}{Theorem}

\begin{document}

\title{Promise and Limitations of Supervised Optimal Transport-Based Graph Summarization via Information Theoretic Measures} 
\author{Sepideh Neshatfar, Abram Magner,   Salimeh Yasaei Sekeh}%

\date{}
\maketitle

\begin{abstract}
Graph summarization is the problem of producing smaller graph
representations of an input graph dataset, in such a way that
the smaller \emph{compressed}
graphs capture relevant structural information
for downstream tasks. One graph summarization
method recently proposed by~\cite{NIPS2019_9014}, formulates an optimal transport-based framework that allows prior information about node, edge, and attribute importance (never defined in that work) to be incorporated
into the graph summarization process. 
However, very little is known about the statistical properties of this framework.  To elucidate this question, we consider the problem of supervised graph summarization, wherein by using information theoretic measures we seek to preserve relevant information about a class label.
To gain a theoretical perspective on the supervised summarization problem itself, we first formulate it
in terms of maximizing the Shannon mutual information between the
summarized graph and the class label.  We show an NP-hardness of approximation result for this problem, thereby constraining what one should expect from proposed solutions.  We then propose a summarization method that incorporates mutual information estimates between random variables associated with sample graphs and class labels into
the optimal transport compression framework. 
We empirically show performance improvements over previous works in terms of classification accuracy and time on synthetic and certain real datasets.
We also theoretically explore the limitations of the optimal transport approach for the supervised summarization problem and we show that it fails to satisfy a certain desirable information monotonicity property.

\end{abstract}

\section{Introduction}
Machine learning involving graphs has a wide range of applications in artificial
intelligence~\cite{scarselli2008graph,dessi2020ai}, network analysis, and biological interactions~\cite{han2019gcn,chen2020gcn}. Graph
classification problems use the network structure of the underlying
data to improve predictive decision outcomes. However, graph datasets are often enormous, and
the algorithms used to extract relevant information from graphs are
frequently computationally expensive. Graph summarization addresses
these scalability issues by computing reduced representations of
graph datasets while retaining relevant information.
As with numerous other problems in machine learning, the precise
meaning of ``reduced representation'' does not have one single
mathematical definition, and there is no single objective
function being optimized. There are thus various approaches to
this problem. For a survey, see~\cite{graph-summarization-survey}.
The particular type of approach of interest in this paper takes as input
a dataset of graphs and a number $k$ and, for each graph $G$ in the dataset, outputs a subgraph $H \subseteq G$ induced by $k$ vertices.

Optimal transport (OT), the general problem of moving one distribution of mass to another as efficiently as possible, has been used in many recent graph-related problems, such as graph matching via the Gromov-Wasserstein distance~\cite{gromov-wasserstein}. One recent approach to the graph summarization problem allows for the incorporation of
user-engineered prior
information is the Optimal Transport based Compression (OTC) 
approach
of~\cite{NIPS2019_9014}.
Their approach is
as follows: a graph $G$, a target number $k$ of vertices,
a probability distribution $\rho_0$ on the vertices of $G$, and a cost function $c:E(G)\to\R$, $E(G)$ denotes the set of edges of $G$, are given as input. Their method computes a probability distribution $\rho_1$ on the vertices of $G$ that has
minimal Wasserstein distance to $\rho_0$, under the constraint that the number of vertices in the support
of $\rho_1$ is at most $k$. Here, the Wasserstein distance is defined with respect to the cost function $c$ (for details, see Section~\ref{sec:preliminaries}). The output subgraph $H$ is
the one induced by the vertices in the support of $\rho_1$.

The authors state that prior information can be incorporated into the method via appropriately
choosing $\rho_0$ and $c$, but in the prior work, this ``prior information'' is not learned, and $\rho_0$ and $c$
are set heuristically, despite the fact that the summarized graphs are from supervised graph learning datasets.  That is, the information about class labels is not used in the summarization framework.  That work does not provide formal justification or precise intuition regarding the conditions under which such an approach is likely to perform well as a graph summarization method.

{\bf The goals of the present work:}
With this context, we can now state the goals of our work.
\begin{itemize}
    \item
        \emph{Give a natural formalization of supervised graph summarization and explore limitations on its computational tractability.}  This is important in order to understand what problem we are trying to solve
        and how well we should expect any method with a reasonable running time to perform.
    \item
        \emph{Adapt the optimal transport summarization framework to incorporate information about class labels.}
        This is reasonable in light of the fact that the OTC framework was evaluated in the original paper on
        supervised graph learning datasets.
    \item
        \emph{Provide theoretical and empirical intuition about the strengths and limitations of the optimal transport approach for supervised graph summarization.}  This is important in order to advance our
        understanding of the relative merits of and inductive biases built into different graph summarization approaches.
\end{itemize}

{\bf Contributions:} We summarize our contributions in this paper as follows.
Toward providing a theory for the limitations of supervised graph summarization methods and thereby solving our first goal stated above, we propose a natural information-theoretic~\cite{TC,YOH} objective function for the task of supervised graph summarization and show that it is NP-hard to optimize, even approximately (we state a formal version of this in Section~\ref{sec:ot-supervised}).  The significance of this is as follows: it tells us that 
the goals of graph summarization methods must be substantially more modest than optimizing our objective function.

 Toward achieving our second goal above, in Section~\ref{sec:ot-supervised}, we propose a novel supervised summarization algorithm based on Optimal Transport that estimates principled values for the parameters from input data.  We show that it empirically surpasses the state-of-the-art performance (including the performance of the specific method proposed in OTC) on selected real and synthetic datasets. The novelty of the summarization algorithm is that we set our optimal transport parameters in terms of node attributes' and edge indicators' mutual information with class variables. 
 We then experimentally show in Section~\ref{sec:experiments} that it outperforms the baseline (no compression) and the unsupervised OTC method in terms of post-compression classification time and test accuracy for fixed compression ratios, despite all such methods suffering from fundamental limitations that we describe next.


In Section~\ref{sec:ot-limitations}, we then explore the limitations of the optimal-transport-based approach, both theoretically and empirically, toward achieving our third goal stated above. 
Specifically, we formulate a notion of information monotonicity of an optimal transport parameter pair with respect to a data distribution. This is the desirable property that means the flow cost decreases monotonically as the mutual information of the resulting summarized graph data with the corresponding class labels increases. This means that optimizing flow cost increases class label information (but may not optimize it). We show that any optimal transport parameter pair satisfying natural properties \emph{fails} to exhibit information monotonicity for at least some data distributions.

\section{Limitations of our approach}
Our framework for supervised graph summarization assumes that all graphs in the input dataset are defined on a 
common vertex set, so that we may straightforwardly speak of the mutual information between edge indicators
and class labels.  Many real datasets satisfy this constraint, and we consider some such in Section~\ref{sec:experiments}.  However, it would be desirable to extend our framework to the case where
graphs may be defined on different vertex sets (as is the case in, for example, many chemistry applications).

Our adaptation of the optimal transport compression framework to the supervised setting, as well as our
study of its limitations, take a local information approach, wherein parameters of nodes/edges are set
on the basis of only their own information content about class labels.  This approach may be extended
to consider more global information, but there is practical reason to take a more local approach (see our discussion below the proof of Theorem~\ref{thm:monotonicity-violation}.

\section{Problem formulation}
{\bf Supervised Graph Summarization} In this section, we give a natural formulation for the supervised graph summarization problem in order
to explore its fundamental limitations.  We assume the following supervised learning context:
a probability distribution $\D$ over tuples $(G, X, C)$ is fixed by nature and unknown to us (as
is the standard assumption in formalizations of supervised learning).  
Here, the graphs $G$ are defined on a single common set $V$ of vertices with cardinality $|V|=n$, $X$ is a $|V|\times d$
matrix whose rows are $d$-dimensional feature vectors corresponding
to the vertices of $G$, and $C$ is a class label coming from some
fixed set $\CC$. A dataset $\{(G_i, X_i, C_i)\}_{i=1}^m
\sim \D^m$ consisting of $m$ independent
and identically distributed (iid) samples from $\D$ is presented
to us.
We fix a target compression ratio $\kappa \in (0, 1)$, which will be the ratio of the number of vertices in a summarized graph to the number in the original graph.
 Our task is, given sample $\{(G_i, X_i, C_i)\}_{i=1}^m$ but no knowledge of distribution $\D$, to select a  subset $H \subseteq V$ of vertices
satisfying the following:
\begin{equation}
     H
     = \argmax_{U \subseteq V, |U| \leq \kappa |V|} I((G[U], X[U]); C),
     \label{expr:Infomax}
\end{equation}
where $(G, X, C)$ is a generic sample from $\D$, and
$G[U]$ is the subgraph of $G$ induced by the vertices in $U$,
and $X[U]$ is the matrix of corresponding feature vectors.
The function $I(\cdot; \cdot)$ is the \emph{Shannon mutual information (MI)}~\cite{coverthomas},
defined as follows: for any random variables $X$ and $Y$ on a common
probability space,
\begin{align}\label{MI}
 I(X; Y)
 = \E_{P(X,Y)}\left[\log \frac{P(X, Y)}{P(X)P(Y)}\right],
\end{align}
where the expectation is taken with respect to the joint distribution
of $X$ and $Y$.  Note that a tuple of jointly distributed random variables, such as $(G[U], X[U])$, is itself a 
random variable, so the expression in (\ref{expr:Infomax}) is well-defined.  Specifically, applying the
definition (\ref{MI}) to (\ref{expr:Infomax}), we get the following expression:
\begin{align}
    &I((G[U], X[U]); C)  =  \E_{(G, X, C) \sim \D}\left[ \log \frac{P_{\D}(G[U], X[U], C)}{P_{\D}(G[U], X[U])P_{\D}(C)}  \right].
\end{align}


Going back to (\ref{expr:Infomax}), intuitively, $H$ is a subset of nodes of $G$ with size at most $\kappa|V|$ whose induced
attributed subgraph has maximum MI with the class label $C$, where the maximization is over
all subsets of nodes with cardinality at most $\kappa|V|$.
To justify (\ref{expr:Infomax}), as we observe in (\ref{MI}), the MI measures the nonlinear dependency between input graphs and their class labels. 
Thus, an optimal subset of vertices that maximizes this measure should preserve classification performance.
This principle of mutual information maximization is at the core of existing approaches to other problems,
such as the information bottleneck objective in supervised clustering~\cite{TishbyInformationBottleneck}.

While this is a very natural formulation of supervised graph
summarization, we immediately run into computational complexity
difficulties, as our first main result, Theorem~\ref{thm:hardness} below, shows.  We first stress
that in reality and in our problem formulation, the summarizer does not know the data-generating distribution
$\D$ -- anything the summarizer uses about it must be learned from data.  We will show that \emph{even if we were given $\D$ (in the sense of being able to query its probability
density function)}, the MI maximization problem would still be computationally hard.

To state the result formally, we define the decision version
of the problem of finding a maximizing $H$ in (\ref{expr:Infomax}) with knowledge of $\D$ as follows:
given a number of vertices $n$, $k \leq n$, query access to $\D$, and a number $\gamma$,
output $1$ if there exists
a subset $H \subseteq [n]$ of vertices with $|H| = k$ and $I((G[H], X[H]); C) \geq \gamma$.
Output $0$ otherwise.

\begin{theorem}[Supervised graph summarization hardness]
    \label{thm:hardness}
    The decision version of the problem of finding a maximizing $H$ in (\ref{expr:Infomax}) with
    knowledge of the data distribution $\D$ is NP-hard.
    
\end{theorem}
{\it Proof:}
We prove this via a reduction from the decision version of the max-clique problem to ours. 
Given a graph $G$ and a number $k \leq n$, the task is to 
determine whether or not $G$ has a $k$-clique. We will do this by constructing an instance of the 
decision version of the supervised
graph summarization problem above, with knowledge of the data-generating distribution.
This entails specifying $\D$ and $\gamma$.

To this end, we first specify $\D$.
We let the class label be $C\sim \Bernoulli(1/2)$.
We construct a complete graph $G'$ on the same node set
as $G$. We assign an edge event $E_{e}$ to each (undirected) edge $e = \{v, w\}$ in $G'$ as follows: if $e$ is in $G$, then we set
    $E_{v,w}$ to be $E_{v,w} \sim \Bernoulli(p)$, where $p = C/2$, independent of any other
edge weight. If, on the other hand, $e$ is not present in $G$, then we set $E_{e}$ to be $0$.
We set all node features to $1$. We next specify the input $\gamma$.
With the distribution $\D$ constructed above, we have that the mutual
information between the subgraph $H$ induced by any set of $k$ nodes in $G'$ is given by  
\begin{align}
     I((H, X[H]); C)
     = I(H; C)
     = \sum_{v\neq w \in V[H]} I(E_{v,w}; C),
\end{align}
where $V[H]$ is the set of the nodes on subgraph $H$.  Setting $D = h(1/4) - h(1/2)/2$, where
$h(x) = -x\log(x) - (1-x)\log(1-x)$, this is $D\cdot {k\choose 2}$ if and only if $H$ is a $k$-clique in $G$, and it is strictly less if there are edges absent from $H$ (i.e., if $H$ is \emph{not} a $k$-clique). 
We thus set $\gamma$ to be $D\cdot {k\choose 2}$.

With $\D$, $n = |V(G)|$, $k$, and $\gamma$ specified, we use a solution to the decision version of the supervised graph summarization problem.  If its output is $1$, we output $1$.  Otherwise, we output $0$.  As we have
argued above, our output is $1$ if and only if $G$ contained a $k$-clique, so that we have solved the decision
version of the max-clique problem.
This completes the proof.
\qed

In fact, the reduction that we exhibited yields a stronger result:
our approximation problem is, in the worst case, NP-hard to approximate
within a constant factor of the optimum. This follows from the
NP-hardness of approximation of the max clique problem. Thus, in the search for practical approaches to graph summarization
in a supervised setting, we must be more modest in our expectations, and we do not solve the above optimization problem, though it remains the intuitive motivator of our approach. In
Section~\ref{sec:ot-supervised}, we describe our approach via optimal
transport and the framework of OTC.

\section{Graph Summarization via OT}
\label{sec:ot-supervised}
In this section, we review the optimal transport summarization framework from \cite{NIPS2019_9014}, then propose our supervised method via information-theoretic measures.
\subsection{Preliminaries: OT on graphs}
\label{sec:preliminaries}
We first describe the framework of optimal transport on graphs.
Fix a graph $G$ on $n$ vertices $[n] = \{1, 2, ..., n\}$ with
edge set $E(G)$. We let $\widehat{E}(G)$ denote the \emph{directed}
edge set of $G$, given by
\begin{align}
     \widehat{E}(G)
    = \{ (v, w) ~|~ \{v, w\} \in E(G) \}.
\end{align}
We define the \emph{signed incidence matrix} $F$ of $G$ as follows:
$F$ is indexed by the set $\widehat{E}(G) \times V(G)$, and if $(v, w) \in \widehat{E}(G)$,
then $F((v, w), w) = 1$ and $(v, w), v) = -1$, and $0$ otherwise.

Let $\rho_0, \rho_1$ be two probability distributions on  $[n]$,
viewed as vectors in $\R^n$. Let $c:E(G)\to\R$ be an arbitrary \emph{cost function}. A \emph{flow} on $G$ is a function $J:\widehat{E}(G)\to [0, \infty)$.  The \emph{cost} of a flow is defined:
\begin{align}
    W(J) = \sum_{e \in \widehat{E}(G)} c(e) J(e).
\end{align}
The \emph{result} of a flow $J$ with initial distribution $\rho_0$
is defined by $ R(J) = \rho_0 + F\cdot J$, where the product $F\cdot J$ is simply the standard
matrix-vector product resulting in a column vector if size $|V|$. 
Finally, the set of flows from $\rho_0$ with the result $\rho_1$
is denoted by $\F_G(\rho_0, \rho_1)$. Next we define
the Wasserstein distance between $\rho_0$ and $\rho_1$ which is the minimum cost of any flow transporting
$\rho_0$ to $\rho_1$ on $G$ as follows:
\begin{align}
    W_{G}(\rho_0, \rho_1)
    = \inf_{J \in \F_G(\rho_0, \rho_1)} W(J).
\end{align}

The method proposed in~\cite{NIPS2019_9014} takes $G$, 
$\rho_0$, $c$, and a compression ratio $\kappa$ as input and
finds a $\rho_1$ with support size $\leq \kappa |V(G)|$ with minimal
Wasserstein distance to $\rho_0$.  The summarizing graph $H$ is
then the one induced by the vertices in the support of $\rho_1$.
The selling point of their method is that prior information about 
``node/edge importance'' can be encoded in $\rho_0$ and $c$.  They do \emph{not} formally define any importance measures.
They perform experiments in which $\rho_0$ assigns a probability to
node $v$ proportional to its degree (so $\rho_0$ is the stationary
distribution of the random walk on $G$, if it is ergodic)
and $c$ assigns a higher cost to edges connecting nodes with different attributes (all of their experiments are on graphs with categorical attributes).  Unlike the framework that we develop in the present paper, their framework is not a priori supervised, though their empirical evaluation is on supervised graph learning datasets.  Furthermore, no justification is given for their choice of $\rho_0$
and $c$. In addition, their framework ignores dataset-level statistics (since their setting of $\rho_0$ and $c$ is entirely dependent on the individual graph being compressed) and ignores
the absence of edges, which can be informative.

\subsection{Our method: supervised OT via information theoretic measures}
\label{sec:our-method}
In this section, we extend the above framework to the supervised setting. The core idea is to use the training 
dataset to estimate measures of informativeness of edge presence/absence and node features about the class label. 
We then use these to construct the initial distribution and cost function, which can then be used in the framework 
of the previous work applied to the complete graph on $n$ nodes to find a subset $H$ of nodes such that the 
subgraphs induced by $H$ in the test set are informative about the class label while 
being a small subset of the original vertex set. 
We run the OTC algorithm on the complete graph
because we wish to take into account the informativeness of the presence/absence of edges.  This informativeness is a statistical property of the data-generating distribution, as reflected in the training dataset; it is \emph{not} a property reflected in a single sample graph.  The method proposed in the prior work does not explicitly incorporate dataset-level statistics, and so does not directly incorporate information about the presence or absence of edges about the class label. 

We define the initial distribution $\rho_0$ by
\begin{align}\label{proposed:rho}
    \rho_0(v) 
    = \frac{I(X_v; C)}{\sum_{w \in [n]} I(X_w; C)},
\end{align}
where $I(X_v; C)$ is the MI between the attributes of node $v$ and the class label $C$, as defined in (\ref{MI}). Next, we propose the following cost function $c(\cdot, \cdot)$:
\begin{align}
    &c(v,w) = D_{KL}(E_{v,w} ~|~ C=0 ~\|~ E_{v,w} ~|~ C=1) + R_{v,w},\;
    \label{expr:cost-definition}
\end{align}
\begin{align}\label{eq:1}
\;\hbox{where }
    R_{v,w} = I(X_v; C ~|~ X_w) + I(X_w; C ~|~ X_v),
\end{align}
and $D_{KL}(A \| B)$ denotes the $KL$-divergence~\cite{coverthomas} from a random variable $A$ to a random variable $B$ (See Algorithm 1).  More precisely, $D_{KL}$ is a functional of pairs of distributions.  In our case,
the distributions are, respectively, that of $E_{v,w}$ conditioned on $C=0$ and that of $E_{v,w}$ conditioned
on $C=1$.
The intuition for this definition of $\rho_0$ is that we consider nodes to be important if their features carry a significant amount of information about the class label.
The intuition behind the definition of cost $c$ in (\ref{expr:cost-definition}) is as follows:
we want a large amount of flow across an edge
if and only if the edge itself is not too informative about
$C$, and the feature of one of the vertices incident on the 
edge is not too informative about $C$, conditioned on the value of the other vertex's feature (in other words, the vertices contain redundant information about $C$).  In this case, we want the edge cost to be small.

Running the optimal transport compression (OTC) routine 
on the complete graph $K_n$ with
$\rho_0, c, \kappa$ as parameters results in a subset $H$ of nodes of cardinality (approximately) $\kappa n$.  To compress a graph $G$
in the test set, we simply take the subgraph of $G$ induced by
$H$.

Crucial to the performance of our method is the ability to estimate
the mutual information and KL divergence quantities in the expressions for $\rho_0$ and $c$ directly without estimating the distributions~\cite{SalimehEntropy2018}.  
We chose EDGE~\cite{noshad2019scalable} because it has been shown that it achieves optimal computational complexity  $O(N)$ where $N$ is the sample size. This is significantly faster than its plug-in competitors~\cite{kraskov2004estimating,moon2017ensemble}. It is proved that in addition to fast implementation, EDGE has the optimal parametric mean squared error (MSE) rate of $O(1/N)$ under a specific condition. 
The KL divergences are estimated via the plug-in method.
In Section~\ref{sec:experiments}, we evaluate our method
on synthetic and generated graphs from real datasets.  We show that it empirically significantly outperforms the previous work on all of the considered datasets in terms of running time and, on some datasets, in terms of test accuracy.




\begin{algorithm2e}[th]   
    \caption{Supervised Optimal Transport Compression}  \label{algg1}
    \KwData{Size $n$ of graphs, target subgraph size $k$, dataset $D = \{(G_j, {X}_j, C_j)\}_{j=1}^m$, where $G_j$ is an $n\times n$ adjacency matrix, ${X}_j \in \R^{n\times d}$ is a matrix of vertex features, and $C_j \in \{0, 1\}$ is a class label}
    \KwResult{A subset $S \subseteq [n]$ of size $k$}
    
    \Comment{Compute initial distribution $\rho_0$ on nodes.}
    \For{$v=1$ to $n$}{
        \Comment{Compute the MI between the features of node $j$ and the class label.}
Set $\rho_0(v) 
 = \frac{I(X_v; C)}{\sum_{w \in V} I(X_w; C)}$\;
    }
Set $s = \sum_{j=1}^n \rho_0(j)$\;
    \For{$j=1$ to $n$}{
Set $\rho_0(j) = \rho_0(j) / s$\;
    }
    
    \Comment{Compute $n\times n$ cost matrix $cost$.}
    \For{$v \neq w$}{
Set $c(v,w) = D_{KL}(E_{v,w} ~|~ C=0 ~\|~ E_{v,w} ~|~ C=1) + (\ref{eq:1})$
    }

    \Comment{ Pass the complete graph, the initial distribution, the edge cost function, and the
    target size $k$ to the OTC routine. }
Set $S = OTCompress(K_n, \rho_0, cost, k)$\;
    \Return{$S$}\;
\end{algorithm2e}

\section{Limitations of the OT approach}
\label{sec:ot-limitations}
In this section, we explore the limitations of the optimal transport approach for the purposes of maximizing the MI between attributed graphs $(G, X)$ and a class label $C$.

To solve this optimization problem via OT we require initial distribution $\rho_0$ and cost function $c$.
Here, we first specify an abstraction of the kinds of initial distribution and cost
function that we propose in (\ref{proposed:rho}) and (\ref{expr:cost-definition}).

\begin{definition}[Data distribution-dependent parameters]
    Fix a graph size $n$. An edge cost function is a function $c$ from
    the joint distribution of $(G, X, C)$ (call this the \textbf{data distribution}) and pairs of vertices
    (i.e., numbers in $[n]$) to non-negative real numbers or $\infty$.
    If the data distribution is given, then $c$ is just a mapping
    from pairs of vertices to $[0, \infty]$.
     
    Similarly, an initial distribution map $L$ is a mapping from
    the data distribution to probability distributions on $[n]$
    (which are meant to be values for $\rho_0$).
    If the data distribution is given, then $\rho_0$ is a function
    mapping vertices to probabilities (which must sum to $1$).
\end{definition}

Below we define a property of
OT parameter pairs called \emph{information monotonicity}, which guarantees that using a pair with this property, flow cost decreases as the MI between the resulting compressed subgraph and the class label increases.  This is a desirable property for parameter pairs, because the optimal transport compression algorithm
in~\cite{NIPS2019_9014} finds an approximately optimal flow iteratively.  Hence, if information monotonicity is violated, then while the cost of the flow may decrease between steps, the MI may also decrease.  We will study to what extent OT parameter pairs satisfy this monotonicity property.

\begin{definition}[Information monotonicity]
    An OT parameter pair $(c, L)$ is said to be \emph{information monotone} for a data distribution $(G, X, C)$ if, for any
    two flows $f^{(0)}, f^{(1)}$ on $G$ resulting in subgraphs
        $H^{(0)}$, $H^{(1)}$, respectively, we have
        \begin{align}
     &W(f^{(0)}) \leq W(f^{(1)})
     \iff  
     I((H^{(0)}, X_{H^{(0)}}); C) \geq I((H^{(1)}, X_{H^{(1)}}); C). \nonumber
     \end{align}
\end{definition}
We have the following theorem regarding non-monotonicity of
optimal transport solutions with respect to mutual information, under natural ``local'' assumptions
on the initial distribution and edge cost function maps. 
\begin{theorem}[Information monotonicity violations in the OTC approach]
    \label{thm:monotonicity-violation}
    Let $c(\cdot, \cdot)$ be a cost function satisfying the
    following properties:
   
          $\bullet$ $c(u, v)$ is strictly monotone increasing with the mutual
            information between $E_{u,v}$ and $C$ and is $0$
            when the two are independent.\\            
    Let $L$ be an initial distribution map satisfying the
    following properties:
    
          $\bullet$   if $X$ is independent of $C$, both unconditionally
            and conditioned on $G$, then $\rho_0$ is the uniform
            distribution.

    Then there exist data distributions for which the parameter pair $(c, L)$ is
    not information-monotone.  In fact, there exist data distributions for which the
    mutual information-maximizing subgraph of a given size has positive MI
    with the class variable, while
    the OTC
    algorithm with parameter pair $(c, L)$ results in a subgraph with MI
    with the class variable arbitrarily close to $0$.

\end{theorem}
{\it Proof:} We will prove this by explicit construction of a data distribution on graphs
    of size $n$ that leads to the claimed monotonicity violation.

    We set all vertex features to be independent of the graph structure and of the class label,
    so they provide no information about either.  In this case, by the assumptions of the
    theorem, $\rho_0$ is the uniform distribution over all $n$ vertices.
    We fix a probability $p$ and generate a class label $C \sim \Bernoulli(p)$.
    Conditioned on $C = b \in \{0, 1\}$, each edge $e = \{v, w\}$ appears with probability
    $q_{e}^{(b)}$, independent of everything else.
    Intuitively, those edges with large $|q_e^{(0)} - q_e^{(1)}|$
    provide more information about the value of $C$ (this is true
    regardless of what the cost function is).  
    We note that in this case, our assumption
    about the cost function translates to assuming that it is a monotone increasing function of
    $|q_e^{(0)} - q_e^{(1)}|$ that is $0$ when $|q_e^{(0)} - q_e^{(1)}|$ is. 
    
    We single out the edge $e_* = \{1, 2\}$ and set
    $|q_{e_*}^{(0)} - q_{e_*}^{(1)}| = const > 0$.  This will be the information-optimal edge.
    For every edge $e = \{v, 1\}$ or $\{v, 2\}$, we set
    $|q_{e}^{(0)} - q_{e}^{(1)}| = const/2$.
    Then, for every remaining vertex pair $e$, we choose $q_e^{(0)} = q_e^{(1)}$
    (in other words, the remaining edges are independent of the class label).
    This implies that $c(e) = 0$.
    This completes the definition
    of the data-generating distribution.

    An optimal flow, over all flows resulting in a distribution $\rho_1$ supported
    on $2$ vertices is as follows: for some vertex pair $\{v, w\}$ such that $v, w \notin \{1, 2\}$,
    for all vertices $u$, the flow sends all $1/n$ of the initial mass on $u$ to $v$ or $w$.
    Since no mass is transported through $e_*$, the cost of this flow is
    $2\cdot (const/2)/n = const/n$.
    Furthermore, the resulting induced subgraph $H$ on vertices $\{v, w\}$ has MI
    $I((H, X[H]); C) = 0$.

    On the other hand, any flow that results in the information-optimal edge must pass a total mass
    of $(n-2)/n$ to the vertices in $e_*$, incurring a cost of $(n-2)/n \cdot const/2$,
    which is strictly greater than the cost of an optimal flow.  This completes the proof. 
    \qed
    
Theorem~\ref{thm:monotonicity-violation} suggests the following intuition behind the shortcomings of a
local information approach to applying the optimal transport compression framework for supervised graph summarization: the neighborhood of an edge's incident vertices
substantially impacts whether or not it will be included in the summarized graph.  One can imagine
defining a cost function and initial distribution that take into account more global information.
However, this necessarily increases the computational complexity of computing these parameters.
Since the goal of graph summarization is, in part, to increase computational efficiency, 
it remains to be seen whether a practical balance can be struck between efficiency and relevant information preservation.



\section{Experimental Study}
\label{sec:experiments}

We conducted several experiments to compare the performance of our method with that of the OT approach and baseline (no-compression). We use graph summarization methods including OTC, Ours (Algorithm~\ref{algg1}), and Coarsening(GC)~\cite{jin2020graph} on each graph classification dataset, then use the resulting summarized dataset to train and test a classifier as described later in this section.
We evaluate our work using both real and synthetic datasets as described below. All the datasets have binary class labels. Note that we first identify class-sensitive nodes for all datasets by using the measure $\rho:=\mathbb{E}_{(X_v,C)\sim D}[{P(X_v|C)}/{P(X_v)}]$ where $P(X_v|C)$ and $P(X_v)$ are conditional and marginal density functions, respectively. $\rho$ determines how the distribution of node features changes given class labels. This way we focus on nodes whose features carry more information for class labels to deliver further emphasis on the contribution of both features and edges in graph classification. The node size in Table~\ref{table.2} is after applying $\rho$.  

{\bf Image-based Dataset:} The real vision datasets that we are using here are based on MNIST \cite{deng2012mnist}, CIFAR-10 \cite{Krizhevsky09learningmultiple} and MiniImageNet \cite{vinyals2016matching} datasets. 
Each pixel in an image is assigned to a node in a graph and each image is a graph of attributed nodes where the edges between the nodes are based on their spatial relationships: adjacent pixels are linked by an edge similar to \cite{cnn_graph}. It is notable that the number of nodes for each graph would be tremendous since images are $784\times784$ pixels and we have a huge adjacency matrix for each graph. So, compressing the nodes here would be necessary specifically when considering a lot of non-useful pixels that might even negatively impact the 
classification. 

{\bf Transport-based Dataset:} As a non-vision real dataset, the New York Taxi Data (2010-2013) \cite{illinoisdatabankIDB-9610843} is used to generate an attributed graph dataset with a fixed node size among all samples. In the raw data, there are about 7 million trips by Taxis in New York City (NYC), and 703 graph samples are extracted with the pre-processing as the following. Firstly, invalid trips are eliminated and then the "vendor\_id"s are considered as the labels. So, the trips are divided into two categories based on their vendor ids. In each category, the trips of each day are considered as one single sample, and the nodes, edges, and nodes' attributes are generated for each sample. For generating the nodes, each element of a grid over the NYC map is assigned to a node. So, each node in each sample shows a neighborhood on the map in a day. The attributes of these nodes are calculated as the summation of the time that taxis spend on a trip when they pick up and drop off people only in that neighborhood. There is an edge between two nodes if there was at least a trip on that day between those two nodes.

{\bf Synthetic Dataset:} The invoked synthetic data has class-label-dependent node features and edges. The feature of node $j$ is generated randomly from a Gaussian distribution with standard deviations and means $j/|V|\times5$ for class $0$ and $j/|V|\times4$ for class $1$. Edges are randomly added to the graphs using the probability assigned to the class label. 

{\bf Experimental design:}
For the classification accuracy experiments, for each dataset and
method, we train the summarization method on a training set, then apply the summarization method with the learned parameters to a test set.
We then use the resulting summarized dataset as input to a classification pipeline: this pipeline is trained/tested using
$5$-fold cross-validation. We report the average and standard deviation of the test accuracy over 5 trials for each compression and classifier. 
We also report the running time of the training (when applicable)
and testing phase of the compression time, as well as the
running time of the classification pipeline on the summarized
dataset. We note that the latter running time is important
because one of the central purposes of graph summarization is
to make subsequent tasks, such as classification, more computationally efficient while not sacrificing too much accuracy. These statistics are \emph{not} reported in prior work~\cite{NIPS2019_9014}.
The same experiment is performed with multiple compression
ratios.
For the classification model, we apply a support vector
machine (SVM) with the GraphHopper kernel~\cite{graphhopper}. We use this kernel to handle continuous node attributes.

\begin{table*}[t!]
\caption{This table demonstrates the accuracy of { OTC}, { Ours}, and Graph Coarsening(GC)~\cite{jin2020graph} methods on several real and synthetic datasets for some compression ratios. The methods with the best performance are indicated in bold font in each case. Here Ours+($\rho$) is the experiment to improve "Ours" accuracy. For fairness "Ours" is the result that should be compared against OTC.} 
\centering 
{\fontsize{8.4}{11}\selectfont
\setlength\tabcolsep{0.9pt}

\begin{tabular}{|l|l|c|c|c|c|c|c|c|c|}
\cline{3-10}

\multicolumn{2}{c}{   }&\multicolumn{8}{|c|} {Compression Ratios}  \\

\toprule
                              Dataset & Method & 0.2& 0.3& 0.4& 0.5 & 0.6 & 0.7& 0.8& 1.0 \\\midrule
\multirow{1}{*}{\shortstack[l]{{\bf MNIST}}} &GC& .538$\pm$.066 & .496$\pm$.068 & .492$\pm$.081 & .558$\pm$.082 & .571$\pm$.070 & .542$\pm$.070 & .571$\pm$.081 & -\\
Graphs:300&   OTC&.687$\pm$.113 & .633$\pm$.123 & .7$\pm$.151 & .673$\pm$.139 & .787$\pm$.167 & .76$\pm$.122 & .747$\pm$.128 & -\\
Nodes: 100 &   Ours& \bf{.937$\pm$.061}& \bf{.945$\pm$.078 }& \bf{.939$\pm$.095 }& \bf{.927$\pm$.115 }& \bf{.917$\pm$.118} & \bf{ .883$\pm$.136} & \bf{ .836$\pm$.169} & -\\
Edges: 179.8 &Baseline &-&-&-&-&-&-&-&.793$\pm$.017 \\
\midrule


\multirow{1}{*}{\shortstack[l]{{\bf CIFAR10}}} &GC& .492$\pm$.002 & .492$\pm$.002 & .506$\pm$.006 & .506$\pm$.006 & .493$\pm$.005 & .493$\pm$.005 & .506$\pm$.006& - \\
Graphs: 600  &OTC& {\bf.516$\pm$.021} & .51$\pm$.027 &  .493$\pm$.029 &  .493$\pm$.039 &  .503$\pm$.046 &  .517$\pm$.035 &  {\bf.53$\pm$.027}& - \\
Nodes: 200&Ours&   .501$\pm$.026 &  {\bf .541$\pm$.031} & {\bf  .549$\pm$.038} &     {\bf.549$\pm$.022} &   {\bf.532$\pm$.033} &   .53$\pm$.026 &   .521$\pm$.021& - \\
Edges: $335.15$&Ours+($\rho$)  &  .507$\pm$.031 &    .522$\pm$.033 &    .536$\pm$.032 &    .53$\pm$.034 &    .528$\pm$.030 & \bf{   .536$\pm$.030 }&   .528$\pm$.026& - \\
&  Baseline& - & - & - & - & - & - & - & .53$\pm$.012 \\
\midrule


\multirow{1}{*}{\shortstack[l]{{\bf MiniImageNet}}} &GC& .493$\pm$.003 & .504$\pm$.006 & .493$\pm$.003 & .493$\pm$.003 & .557$\pm$.003 & .504$\pm$.006 & .504$\pm$.006& - \\
Graphs: 1000  &OTC& .573$\pm$.027 & {\bf.576$\pm$.022 }&  {\bf .583$\pm$.020} &  .564$\pm$.037 &  .559$\pm$.041 &  .569$\pm$.040 &  {\bf .571$\pm$.029}& - \\
Nodes: 250&Ours&  .565$\pm$.039 &   .562$\pm$.043 &  .568$\pm$.042 &     .561$\pm$.041 &   {\bf .569$\pm$.038} &   {\bf .569$\pm$.039} &   .567$\pm$.043& - \\
Edges: 111.06&Ours+($\rho$)   &   \bf{ .58$\pm$.029} &    .57$\pm$.038 &    .566$\pm$.043 &  \bf{    .566$\pm$.042} &    .569$\pm$.039 &   .568$\pm$.038 &   .57$\pm$.036& - \\
&  Baseline& - & - & - & - & - & - & - & .569$\pm$.028 \\
\midrule

\multirow{1}{*}{\shortstack[l]{{\bf NYC}}} &GC& .524$\pm$.048 & .529$\pm$.077 & .522$\pm$.037 & .520$\pm$.064 & .510$\pm$.088 & .525$\pm$.062 & .5256$\pm$.041&-\\
Graphs: 703 &OTC&  N/A & .874$\pm$.051&  .922$\pm$.031 & {\bf .952$\pm$.02 }&  .918$\pm$.025 &  .888$\pm$.03 & {\bf .918$\pm$.038}&-\\
Nodes: 100 &Ours&  N/A & {\bf  .96$\pm$.017 }& {\bf  .955$\pm$.020} &   .943$\pm$.026 & {\bf  .929$\pm$.031 }& {\bf  .907$\pm$.047} &   .889$\pm$.059&-\\
Edges:: 418.13&  Baseline&- &- &- &-&-&-&-&.871$\pm$.04 \\
\midrule
\multirow{1}{*}{\shortstack[l]{{\bf Synthetic}}} &GC& .625$\pm$.026 & .655$\pm$.053 & .656$\pm$.013 & .708$\pm$.031 & .732$\pm$.035 & .734$\pm$.026 & .763$\pm$.027&-\\
Graphs: 1500 &OTC&   .886$\pm$.039 &   .918$\pm$.016 &   .957$\pm$.011 &  .966$\pm$.013 &   .984$\pm$.005 &  .986$\pm$.004 &  .992$\pm$.003&-\\
Nodes: 100 &Ours& {\bf   .909$\pm$.014 }& {\bf  .950$\pm$.01 }& {\bf  .981$\pm$.005 }& {\bf   .984$\pm$.007} & {\bf  .986$\pm$.014} & {\bf  .994$\pm$.003 }& {\bf  .993$\pm$.004}&-\\
Edges:4017.41 &Baseline &-& -& -& -& -& -& - &.992$\pm$.005\\
\bottomrule                           
\end{tabular}
}
\label{table.2}
\end{table*}


\subsection{Performance Analysis on graph classification datasets}
To evaluate our method and analyze its performance
we compared the test accuracy of our method, OTC, Graph Coarsening (GC)~\cite{jin2020graph}, and baseline (i.e. no compression). Among other methods, we focused on only Graph Coarsening and OTC which already outperforms several existing works. Table~\ref{table.2} summarizes the performances for each compression ratio. As the numbers in bold indicate, our method outperformed the other methods across the datasets on most compression ratios in expectation. Our method outperforms OTC, GC, and baseline on MNIST, NYC, and synthetic datasets for all compression ratios while for CIFAR10 and miniImageNet we could beat OTC on some of the ratios (0.3-0.6 for CIFAR10 and 0.6, 0.7 for miniImageNet).


It is notable that we outperform other methods' test accuracy in half of the real datasets, along with the synthetic dataset, by a large margin. We hypothesize that this is not happening for CIFAR10 and MiniImageNet because our chosen optimal transport parameters only incorporate the local informativeness of each node. In such complex image classification tasks, pixels in isolation are less informative than the global image structure. This may result in our method failing to perform significantly better than the baseline in terms of test accuracy. To improve the accuracy of our method for CIFAR10 and miniImageNet, we have run another experiment using again $\rho$ (called "Ours+$\rho$ on Table~\ref{table.2}). We succeed on $0.7$ ratio for CIFAR10 and $0.2$, $0.5$ ratio for miniImageNet. Note that for the sake of fairness "Ours" is the result that should be compared against OTC as the sensitive nodes are selected only once in OTC.

As observed in the table for most of the examined datasets, the standard deviations are sufficiently large similar to the other well-known graph compression methods-- that we cannot make empirical high-probability statements of superiority over other methods. Here ``baseline'' means a classification performance on a non-summarized graph. 

\begin{figure}[h]
     \centering
\subfloat{
    \includegraphics[width=0.45\columnwidth]{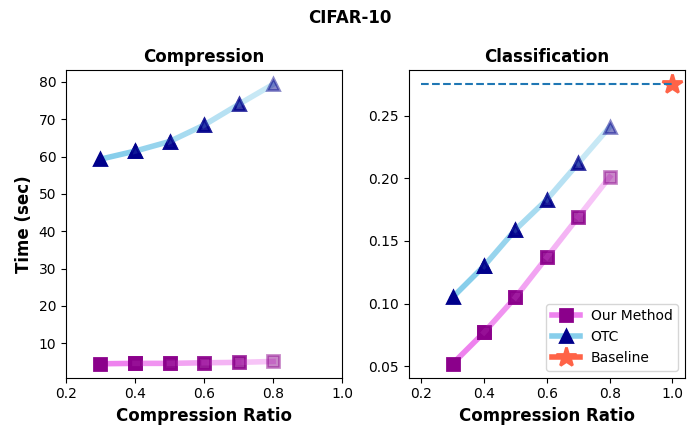}}
    \label{fig:r50_feature_distributions}
\subfloat{
    \includegraphics[width=0.45\columnwidth]{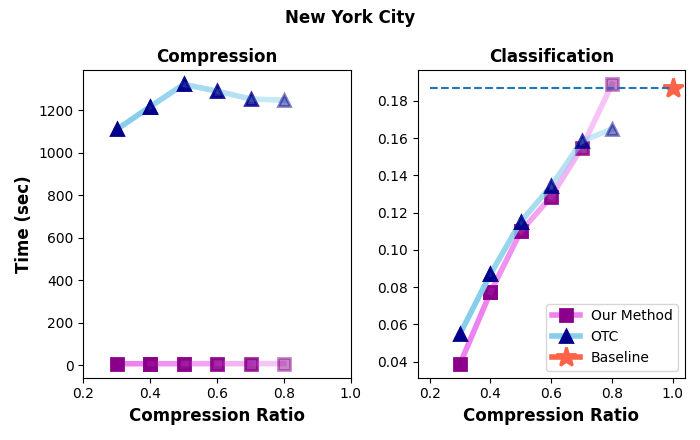}}
    \label{fig:r50_baseline}
\caption{The compression and classification time for both Ours and OTC methods is shown for both CIFAR-10 and NYC. Our method substantially outperforms OTC and baseline (no compression) in both classification and compression times. The classification time for the compression ratio with the best accuracy for our method is less than half of that of the baseline. }
\label{time-experiments}
\end{figure}

Furthermore, classification error is not the only important evaluation metric; training and testing speed is also essential. Our method is uniformly superior to the baseline in terms of these metrics, because the baseline performs the optimal transport summarization on each new graph, whereas in our method this is only done on the training set. Figure~\ref{time-experiments} demonstrates that our method outperforms OTC with a very large margin in terms of compression time. It shows that our method outperforms both OTC and baseline (no compression) in terms of classification times. This is because they find the best-compressed vertex set for each sample individually while we find a single best-compressed vertex set for the whole sample.

\section{Conclusion}
We have introduced a theoretical formulation of supervised graph summarization
in terms of maximization of the mutual information between the class labels and
graphs to be classified.  We showed that the solution to this problem, even approximately, is NP-hard, and so we took a different approach.
In particular, we showed how
the unsupervised optimal transport graph summarization framework of~\cite{NIPS2019_9014} can be adapted -- nontrivially -- to a supervised setting via estimation of information-theoretic measures incorporating both graph structure and node features.  This is in contrast to the parameter settings of the previous work, which only took into account the degree distribution of the graph and compressed graphs in isolation, failing to take into account dataset-wide statistics.  We also elucidated limitations of the optimal transport solution for maximization of the mutual information objective in our summarization framework.\\

{\bf Looking ahead} - 
Further work is necessary to elucidate the precise power/limitations of the optimal transport approach to graph summarization.  For instance, its connections to other methods, such as graph clustering-based methods deserve further examination.
Moreover, our adaptation of the optimal transport method
to the supervised setting requires further development to accommodate graph datasets in which different graphs do not share a common vertex set. 

\section{Acknowledgement}
This work has been partially supported (Sepideh Neshatfar and Salimeh Yasaei Sekeh) by NSF-CAREER 5409260 and (Abram Magner) by NSF-CCF 221232; the findings are those of the authors only and do not represent any position of these funding bodies.

%
%
\renewcommand{\refname}{References}
\bibliographystyle{IEEEbib}

\bibliography{egbib}


\newpage
\section{Appendices}
In this section, further illustrations and analysis of datasets and parameters are provided.

\begin{figure}[h!]
  \centering
    \includegraphics[width=0.7\columnwidth]{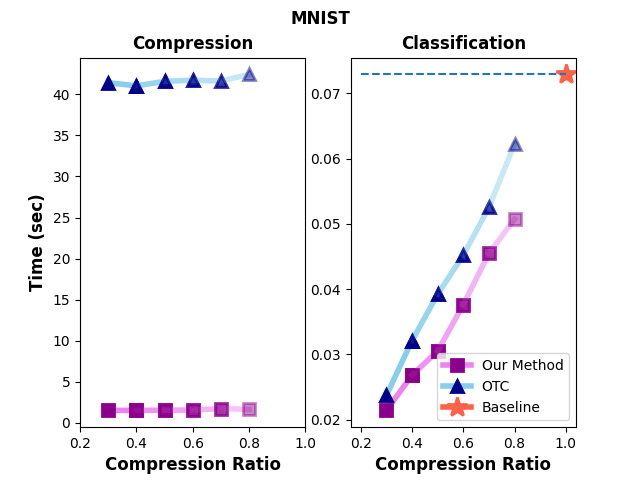}
    \caption{{\it MNIST Compression and Classification time comparison} In the Compression comparison figure on the left, the difference between OTC and our compression time is remarkable on MNIST. Also, for the smallest compression ratio with the best performance the classification time is less than half of baseline classification time. }
    \label{MNIST-runtime}
\end{figure}

\subsection{Further empirical study on $\rho$}
\begin{figure}[h!]
  \centering
    \includegraphics[width=0.5\columnwidth]{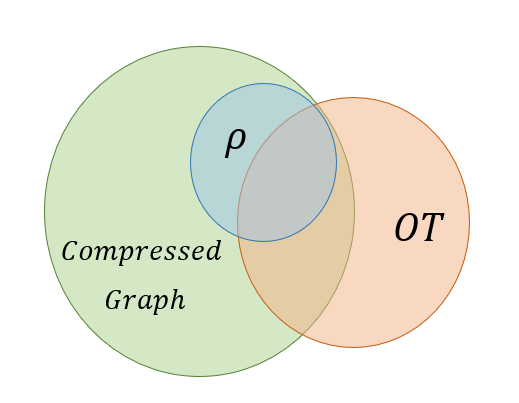}
    \caption{{\it Relation between compressed graph, sensitive nodes, and OT.} The final compressed graph should embrace all sensitive nodes, but might not have some OT without $\rho$. }
    \label{rho-overlap}
\end{figure}

$$\rho:=\mathbb{E}_{(X_V,C)\sim D}\left[\frac{P(X_V|C)}{P(X_V)}\right]$$
where $P(X_V)$ is the marginal probability and $P(X_V|C)$ is the conditional probability of node attributes given class variable i.e. $$P(X_V|C)=\sum_{c}P(X_V|C=c) \pi(c) \;\;\hbox{where}\;\; \pi(c)\;\hbox{is prior probability of class}\; c$$
To clarify what it means to have $\rho$s with different percentages, we should consider their multiplication with the compression ratio as the ratio of the nodes that we tag as sensitive. But these sensitive nodes may or may not overlap with the ones that we compress using optimal transport (OT) in Algorithm 1. The final result of both $\rho$ and OT compressed graph would have the proportion out of nodes equal to the compression ratio. To illustrate this look at Figure~\ref{rho-overlap}.

\begin{table*}[h]
\caption{The table below the performance of our method with different $\rho$ ratios for different compression ratios on CIFAR-10 dataset}
\centering
\scalebox{0.84}
{
\begin{tabular}{@{}llccccccc@{}}
\toprule
                             $\rho$\% & Acc@0.2& Acc@0.3& Acc@0.4& Acc@0.5 & Acc@0.6 & Acc@0.7& Acc@0.8\\\midrule
 20  & {\bf0.507$\pm$0.031 }& {\bf0.522$\pm$0.033} & {\bf0.536$\pm$0.032 }& {\bf0.53$\pm$0.034} & 0.527$\pm$0.034 & 0.523$\pm$0.027 & 0.525$\pm$0.025\\
40  &  0.505$\pm$0.031 & 0.52$\pm$0.034 & 0.531$\pm$0.035 & {\bf0.53$\pm$0.034} &  0.525$\pm$0.031 & 0.528$\pm$0.031 & 0.526$\pm$0.028\\
60  & 0.507$\pm$0.034 & 0.512$\pm$0.040 & 0.523$\pm$0.038 & 0.523$\pm$0.033 & 0.525$\pm$0.030 & 0.534$\pm$0.032 & {\bf 0.528$\pm$0.026}\\
80  & 0.505$\pm$0.035 & 0.506$\pm$0.045 & 0.518$\pm$0.038 & 0.522$\pm$0.032 & {\bf0.528$\pm$0.030} & {\bf0.536$\pm$0.030} & {\bf0.528$\pm$0.026}\\

                            \bottomrule
\end{tabular}}
\label{table.3}
\end{table*}
\newpage
\subsection{NYC Dataset Generation}
\begin{figure}[h!]
  \centering
    \includegraphics[width=0.4\columnwidth]{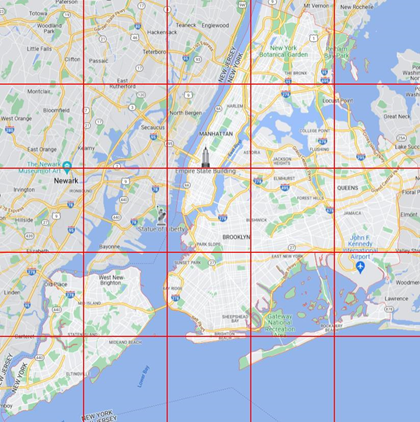}
    \caption{{\it NYC Grid map} Each square of the grid shows the nodes of graphs in NYC dataset. The attributes of the nodes are the total time of the trips inside that region. There is an undirected edge between two Nodes if there was any trip between them. }
    \label{NYC-grid}
\end{figure}

Figure ~\ref{NYC-grid} shows how each area on NYC map is assigned to a node in its graph version.



\end{document}